%% file: root.tex
\documentclass[letterpaper, 10 pt, conference]{ieeeconf}  

\IEEEoverridecommandlockouts                              

\usepackage{amsmath} 
\usepackage{amssymb}  
\usepackage{graphicx}

\usepackage{booktabs}
\usepackage{multirow}
\usepackage[table]{xcolor}

\title{\LARGE \bf
ModalPatch: A Plug-and-Play Module for Robust Multi-Modal 3D Object Detection under Modality Drop
}

\author{ Shuangzhi Li\textsuperscript{\rm 1}, Lei Ma\textsuperscript{\rm 2,1}, and Xingyu Li\textsuperscript{\rm 1} \\
\textsuperscript{\rm 1}University of Alberta, Canada \quad  \textsuperscript{\rm 2}The University of Tokyo, Japan\\
\{shuangzh, xingyu\}@ualberta.ca, ma.lei@acm.org
}

\begin{document}

\maketitle
\thispagestyle{empty}
\pagestyle{empty}

\input{sec/0_abstract}
\input{sec/1_intro}
\input{sec/2_related}
\input{sec/3_method}

\input{sec/4_experim}
\input{sec/5_conclude}

\section*{ACKNOWLEDGMENT}
This work was supported in part by Canada CIFAR AI Chairs Program, the Natural Sciences and Engineering Research Council of Canada, JST CRONOS Grant (No. JPMJCS24K8), JSPS KAKENHI Grant (No.JP21H04877, No.JP23H03372, and No.JP24K02920).

\bibliographystyle{IEEEtran}   
\bibliography{IEEEabrv,refs}

\end{document}

%% file: sec/0_abstract.tex
\begin{abstract}

Multi-modal 3D object detection is pivotal for autonomous driving, integrating complementary sensors like LiDAR and cameras. However, its real-world reliability is challenged by transient data interruptions and missing, where modalities can momentarily drop due to hardware glitches, adverse weather, or occlusions. This poses a critical risk, especially during a simultaneous modality drop, where the vehicle is momentarily blind. To address this problem, we introduce ModalPatch, the first plug-and-play module designed to enable robust detection under arbitrary modality-drop scenarios. Without requiring architectural changes or retraining, ModalPatch can be seamlessly integrated into diverse detection frameworks. Technically, ModalPatch leverages the temporal nature of sensor data for perceptual continuity, using a history-based module to predict and compensate for transiently unavailable features. To improve the fidelity of the predicted features, we further introduce an uncertainty-guided cross-modality fusion strategy that dynamically estimates the reliability of compensated features, suppressing biased signals while reinforcing informative ones. Extensive experiments show that ModalPatch consistently enhances both robustness and accuracy of state-of-the-art 3D object detectors under diverse modality-drop conditions. Code will be available at https://github.com/Castiel-Lee/MM3Det\_MD.

\end{abstract}

%% file: sec/1_intro.tex
\section{INTRODUCTION}

Multi-modal 3D object detection~\cite{yuan2025survey, zhang2025survey} has become a cornerstone in applications such as autonomous driving and robotics. By integrating complementary sensing modalities such as LiDAR and cameras, these systems achieve more accurate and reliable object localization and classification. 
However, despite these benefits, multi-modal systems face critical robustness challenges during real-world deployment. Sensor inputs are often susceptible to failure~\cite{liao2025benchmarking,  bijelic2018benchmark} due to a variety of unpredictable conditions, including hardware malfunction~\cite{chang2023failure, goelles2020fault}, adverse weather~\cite{bijelic2019seeing, bijelic2018benchmark}, and occlusion~\cite{xu2022behind, goelles2020fault}. Moreover, asynchronous sensors may produce misaligned or incomplete data because of differences in sampling frequencies~\cite{caesar2020nuscenes}. These issues result in modality drop (i.e., temporary or partial loss of sensor inputs) that degrade detection performance if not properly handled. All of these real-world challenges point to a fundamental problem in multi-modal 3D object detection:
\begin{center}
\textbf{\textit{How can we ensure robust detection when one or more modalities drop unexpectedly?}}
\end{center}


Existing approaches to this problem~\cite{yan2023cross, wang2024unibev, cha2024robust} have two significant limitations. First, they primarily address a simplified scenario: \textit{dependent} modality-drop, where at least one modality (e.g., LiDAR or camera) is guaranteed to remain available. This overlooks a more critical and realistic scenario: a simultaneous modality drop, where transient issues cause all sensors to lose signal concurrently for short periods. Though brief, such events create moments of total perceptual blindness for an autonomous system. Second, these solutions often require re-designing the core detection architecture or complete model retraining~\cite{wang2024unibev, cha2024robust}, making them resource-intensive, inflexible, and difficult to generalize across different state-of-the-art (SOTA) detectors. 
\begin{figure}[t]
      \centering
      \includegraphics[width=0.45\textwidth]{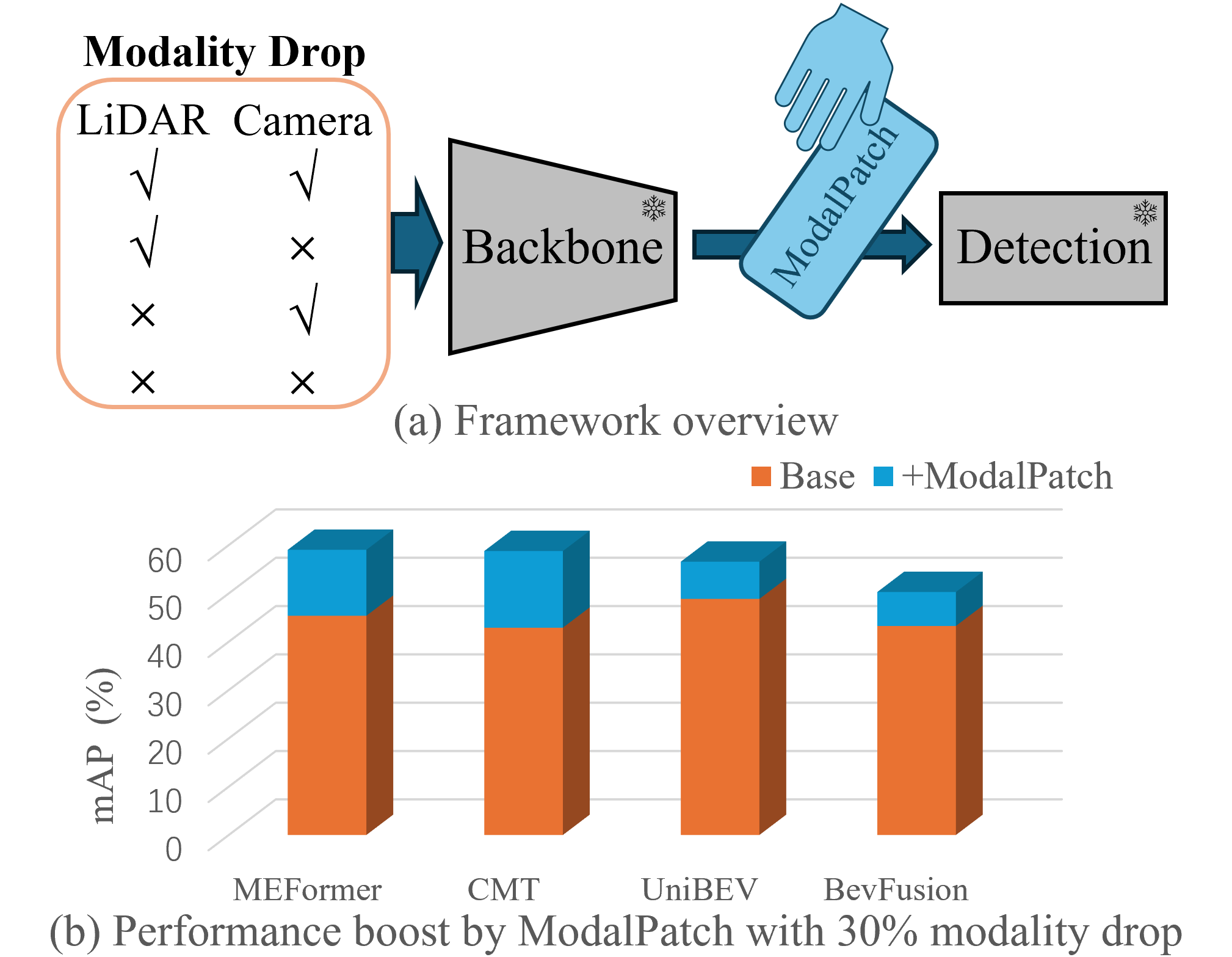}
      \caption{(a) The proposed ModalPatch framework compensates for arbitrary modality drops (LiDAR or camera) and can be seamlessly integrated into existing detectors without retraining them. (b) Performance boost achieved by ModalPatch under the 30\% modality drop rate for various detectors.}
      \label{fig:highlight}
   \end{figure}

To bridge these research gaps, we propose ModalPatch, a lightweight, plug-and-play solution designed to make multi-modal 3D detectors resilient to transient modality drops, regardless of whether the drop is dependent or simultaneous (independent). 
It can be seamlessly integrated into a variety of existing detection frameworks without requiring re-training or architecture modifications. 
Specifically, our key insight is that 3D perception systems in domains like autonomous driving and robotics typically operate on temporal data streams, where sensory inputs evolve continuously over time. ModalPatch exploits this property by maintaining a short-term historical memory of past feature representations, which it uses to predict and compensate for current missing modality data. 
To further improve robustness, ModalPatch incorporates a cross-modality fusion module that promotes interaction between modalities, allowing the system to mitigate the weaknesses of individual sensors. However, predictions based on temporal history may carry noise or bias. To address this, we introduce an uncertainty-aware mechanism in our cross-modality fusion module. Estimating the reliability of both the predicted and live features, this mechanism suppresses unreliable components and amplifies trustworthy ones, yielding more robust detection outcomes. We validate ModalPatch through extensive experiments on multiple SOTA 3D detectors, demonstrating its ability to generalize across architectures and consistently improve performance under diverse modality-drop conditions.

Our contributions are summarized as follows:
\begin{itemize}
    \item We propose ModalPatch, the first plug-and-play solution that can be seamlessly integrated into existing 3D object detection frameworks to handle arbitrary modality drop. 
    \item We leverage the temporal nature of 3D perception by using historical feature memory to predict missing modality features. This serves as an effective and adaptive compensation mechanism in dynamic environments where sensor inputs may drop unexpectedly.
    \item We introduce an uncertainty-guided cross-modality fusion strategy. This mechanism suppresses unreliable signals and reinforces trustworthy ones, improving robustness against both feature bias and noise propagation.
    \item We demonstrate that ModalPatch consistently improves robustness on various SOTA detectors and generalizes well across different 3D detection architectures and a wide range of modality-drop scenarios.
\end{itemize}

%% file: sec/2_related.tex
\section{RELATED WORKS}
\subsection{Multi-modal 3D Object Detection}
Multi-modal 3D object detection~\cite{yuan2025survey, zhang2025survey, song2024robustness} has rapidly advanced by leveraging the strengths of multiple modalities (e.g., LiDAR and camera). Based on feature extraction from heterogeneous modalities, recent methods can be broadly categorized into two main branches: 1) BEV-based feature fusion, where features from different modalities are projected into a shared bird’s-eye-view (BEV) representation to facilitate cross-modal alignment and joint reasoning, as exemplified by BEVFusion~\cite{liu2022bevfusion,liang2022bevfusion} and its extensions~\cite{wang2024unibev, cai2023bevfusion4d}; 2) transformer-based cross-modal interaction, where heterogeneous modality features are fused with attention mechanisms via spatial-aware queries with
unified 3D position embedding~\cite{liu2022petr}, as in Cross-Modal Transformer (CMT)~\cite{yan2023cross}, TransFusion~\cite{bai2022transfusion}, and SparseLIF~\cite{zhang2024sparselif}. While these detectors achieve strong performance under normal conditions, they often experience significant degradation when one or more modalities are missing.
   
\subsection{Multi-modal 3D Object Detection with Modality Drop}

Traditional trials adopt the Kalman filter to compensate for missing inputs by estimating the target's temporal state~\cite{sinopoli2004kalman}.
Recent studies have explored feature compensation for multi-modal 3D object detection: CMT~\cite{yan2023cross} enhances robustness by randomly masking camera inputs during training; UniBEV~\cite{wang2024unibev} introduces unified BEV queries for feature extraction and channel-adaptive fusion; and MEFormer~\cite{cha2024robust} employs dynamic ensembling of detectors trained on different modality subsets to handle both single- and multi-modality inputs. While these approaches represent important progress, they still fall short in addressing the challenges highlighted earlier: they primarily assume dependent modality drop (at least one modality exists) rather than truly independent random drop (no modality input at some time points), and they often require substantial architectural redesign or retraining, limiting their practicality. In contrast, our work introduces a plug-and-play module that can be seamlessly integrated into existing detectors to handle arbitrary modality drops, thereby ensuring robust performance without sacrificing flexibility or generality.

%% file: sec/3_method.tex
\section{METHODOLOGY}
\begin{figure*}[h]
      \centering
      \includegraphics[width=0.90\textwidth]{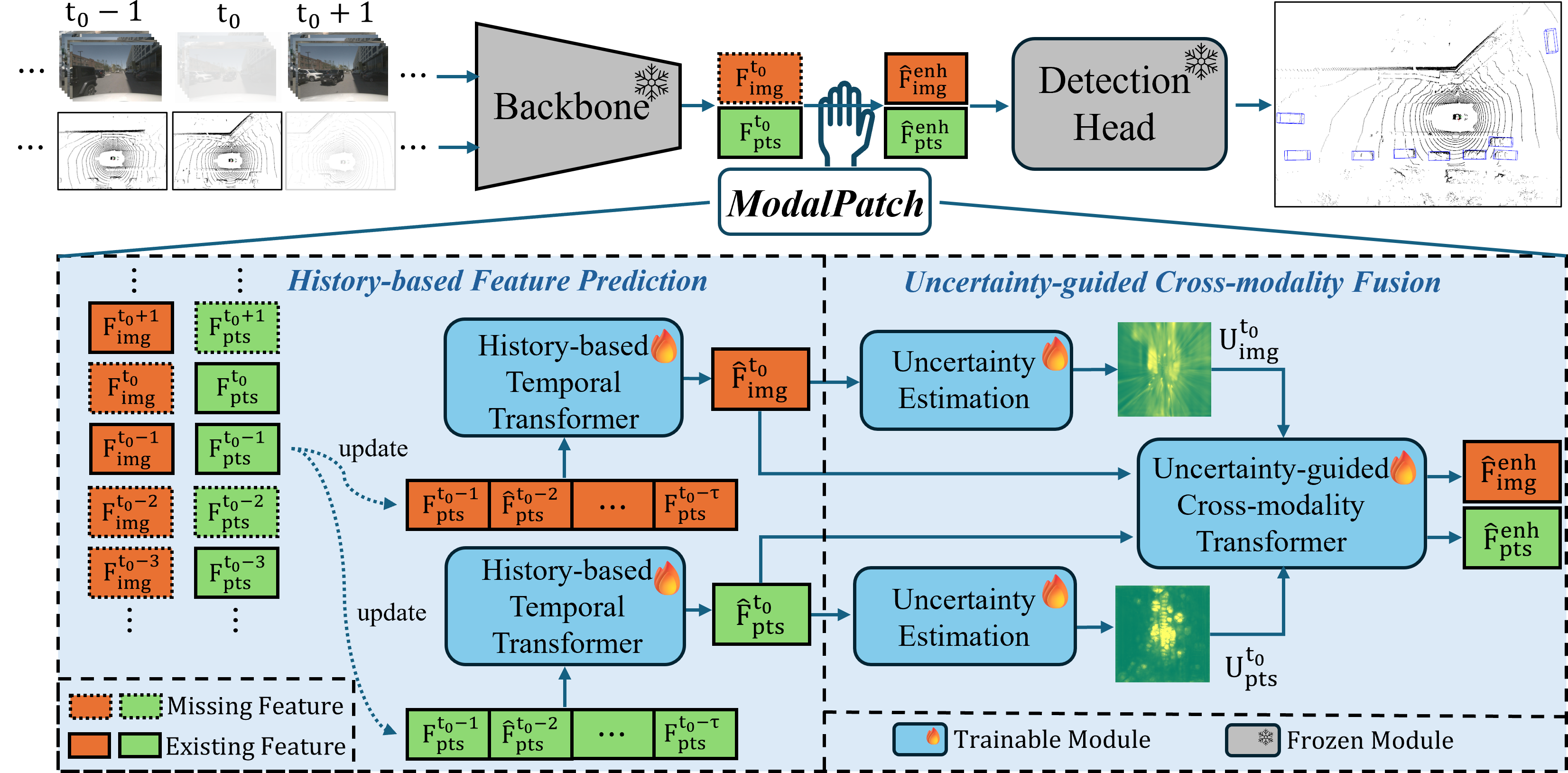}
      \caption{Overview of the proposed ModalPatch module. Given multi-modal inputs (e.g., point clouds by LiDAR and images by camera), features are first extracted by the frozen backbone. To address possible modality-drop scenarios, the plug-and-play ModalPatch introduces two modules: (1) a history-based feature prediction module, which leverages temporal information from past frames to predict current features and provide initial compensation; and (2) an uncertainty-based cross-modality fusion module, which estimates spatial uncertainty and employs cross-modality complementary information to enhance compensated features.}
      \label{fig:framework_overall}
   \end{figure*}

ModalPatch is a lightweight “patch” module that can be seamlessly plugged into diverse detection frameworks for various modality drop scenarios. As demonstrated in Fig.~\ref{fig:highlight}, by simply attaching this module, various 3D detectors can be patched to recover missing modalities — whether single or multiple, thus maintaining robust detection performance in the presence of arbitrary modality loss. 
As shown in Fig.~\ref{fig:framework_overall}, ModalPatch combines (1) \textit{history-based feature prediction}, which leverages temporal continuity to compensate for missing features, and (2) \textit{uncertainty-guided cross-modality fusion}, which improves these features through uncertainty-aware cross-domain complementary refinement. 

\subsection{History-based Feature Prediction (HFP)}
Temporal continuity is inherent in multi-modal 3D object detection and has been extensively explored in recent studies~\cite{cai2023bevfusion4d, zong2023temporal}. Inspired by~\cite{zong2023temporal}, we design a history-based feature prediction module that leverages temporal observations to compensate for missing modalities. Specifically, we model the evolution of features over a sequence of past frames and use the most recent historical memory to predict the current features, which are then employed as substitutes for the missing modality inputs. Moreover, during deployment, the compensated features can serve as an alternative history to ensure continuous compensation and provide relatively stable detection performance over time.

\begin{figure}[h]
      \centering
      \includegraphics[width=0.47\textwidth]{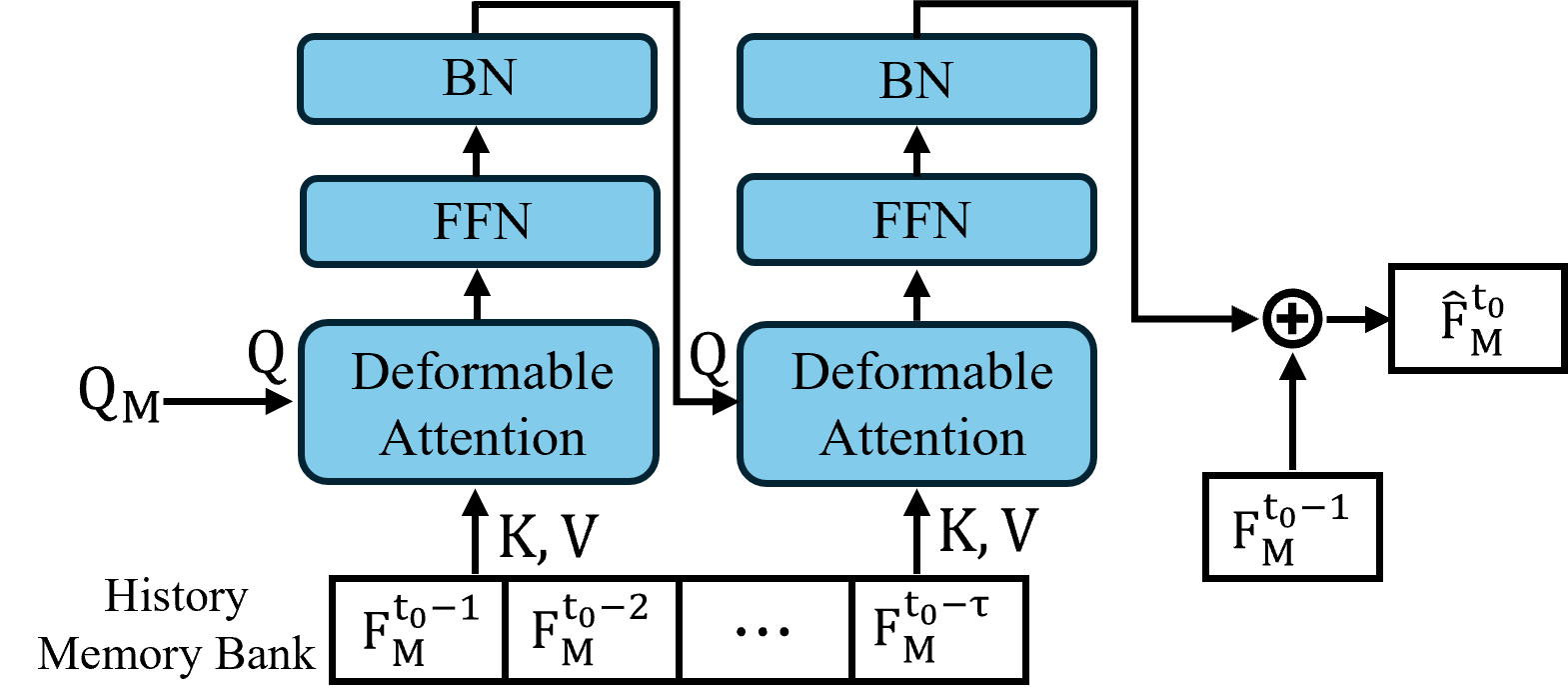}
      \caption{History-based temporal transformer, taking learnable queries and the history memory bank as inputs to generate compensated features. }
      \label{fig:HFP}
   \end{figure}

\noindent \textbf{History-based Temporal Transformer.} 
The input images and point clouds are first processed by their respective backbones to obtain feature maps $F_{M}^{t_0}\in\mathbf{R}^{B \times D_{M} \times H_{M} \times W_{M}}$, where $t_0$ denotes the current time and $\{B, D_{M}, H_{M}, W_{M}\}$ denote the batch size, channel dimension, height, and width of the extracted features for modality $M\in\{\text{img}, \text{pts}\}$. As shown in Fig.~\ref{fig:HFP}, we maintain a memory bank of historical features $\{F_{M}^{t_0-1},F_{M}^{t_0-2},\ldots,F_{M}^{t_0-\tau}\}$ with a temporal length of $\tau$ for each modality $M$ to store historical information. Then we concatenate these history features as $S_{M}$ and use them as the key and value, while the learnable BEV embedding $Q_{M}$ serves as the query to model feature dynamics across time. Specifically, we adopt spatially sensitive deformable attention~\cite{xia2022vision} to adaptively aggregate local context and capture fine-grained temporal dynamics:  
\begin{equation}
\tilde{F}_{M}^{t_0} = \mathrm{DeformAttn}\big(\mathrm{DeformAttn}(Q_{M}, S_{M}), S_{M}\big),
\end{equation}
where $\tilde{F}_{M}^{t_0}$ denotes the compensated temporal dynamics. We then add this dynamic to the most recent historical feature $F_{M}^{t_0-1}$ to obtain the compensated feature for the current time:     
\begin{equation}
\hat{F}_{M}^{t_0} = F_{M}^{t_0-1} + \tilde{F}_{M}^{t_0}.
\end{equation}

Since training datasets collected in autonomous driving and robotics generally provide complete multi-modal inputs, we reasonably assume that modality missing does not occur during training. Thus, the ground-truth feature $F_{M}^{t_0}$ can be used to supervise prediction accuracy of $\hat{F}_{M}^{t_0}$ via an $\ell_2$ loss~\cite{lee2013study}:
\begin{equation}
\mathcal{L}_{\text{TemPred}} = \| \hat{F}_{M}^{t_0} - F_{M}^{t_0} \|_2^2.
\end{equation}

\noindent \textbf{History Memory Update.} 
During training, the history memory bank is updated using the ground-truth features $F_{M}^{t_0-1}$ extracted by the backbone, which ensures accurate temporal supervision. At inference time, for modalities that are available, the bank is updated with the extracted features $F_{M}^{t_0-1}$, while for missing modalities, we instead use the compensated features $\hat{F}_{M}^{t_0-1}$ as substitutes. This strategy guarantees temporal continuity of the memory bank, allowing the model to maintain reliable predictions over consecutive time steps even under arbitrary modality drop.  
  
\subsection{Uncertainty-guided Cross-modality Fusion (UCF)}
Although modality compensation alleviates the problem of missing inputs, each single modality still suffers from inherent limitations~\cite{yuan2025survey,liu2022bevfusion}, such as the restricted field of view and incomplete geometrical cues. To overcome these constraints, we perform cross-modality fusion to leverage complementary information across sensors. However, due to error accumulation in temporal prediction and modality-specific deficiencies mentioned above, the compensated features may contain unignorable biases, which can propagate through the fusion process and degrade detection performance. To mitigate this issue, we design an uncertainty-guided fusion module, which explicitly estimates the reliability of compensated features and adaptively weights their contributions. This mechanism suppresses biased or unreliable signals while reinforcing trustworthy ones, thereby enabling robust cross-modality feature enhancement and improving overall detection accuracy.

\noindent \textbf{Uncertainty Estimation.} 
Uncertainty estimation has been widely studied as an effective approach to assess the reliability of model behavior~\cite{kendall2017uncertainties, wang2025uncertainty}. To mitigate potential cumulative errors and bias in compensated features, we propose to estimate the spatial uncertainty of the compensated features and use it as a reliability measure to adaptively weight their contributions in subsequent cross-modality fusion. 

Concretely, for each spatial location in the compensated feature map $\hat{F}_{M}^{t_0}$, we assume the predicted value follows a Gaussian distribution centered at the corresponding ground-truth feature $F_{M}^{t_0}$, with an unknown variance $\sigma^2_{M}$. To estimate this uncertainty, we employ a lightweight MLP to regress the variance from $\hat{F}_{M}^{t_0}$, producing a variance map:
\begin{equation}
    \sigma^2_{M} = \text{MLP}(\hat{F}_{M}^{t_0})\in\mathbf{R}^{B \times 1 \times H_{M} \times W_{M}}.
\end{equation}
To optimize the MLP, we adopt the negative log-likelihood (NLL) loss~\cite{kumar2019uglli}, which encourages the predicted variances to reflect the reliability of the compensation.
\begin{equation}
    \mathcal{L}_{\text{Uncert}}=\sum_{i=1}^{H_{M}W_{M}} 
\frac{1}{2}(\frac{\lVert \hat{F}_{M,i}^{t_0}-F_{M,i}^{t_0} \rVert_2^2}{\sigma^2_{M,i}}
+ \log\sigma^2_{M,i} + \log(2\pi)),
\end{equation}
where $\sigma^2_{M,i}$ denotes the predicted variance at spatial location $i$. Minimizing this loss encourages the model to assign higher variance to less reliable regions, thereby allowing $\sigma^2_{M}$ to serve as an uncertainty estimate. We then define the uncertainty map $U_{M}^{t_0}:=\sigma_{M}$, which captures the spatial reliability of the compensated features $\hat{F}_{M}^{t_0}$ and 
guide the subsequent cross-modality fusion to suppress uncertain regions while emphasizing more confident signals.

\begin{figure}[t]
      \centering
      \includegraphics[width=0.3\textwidth]{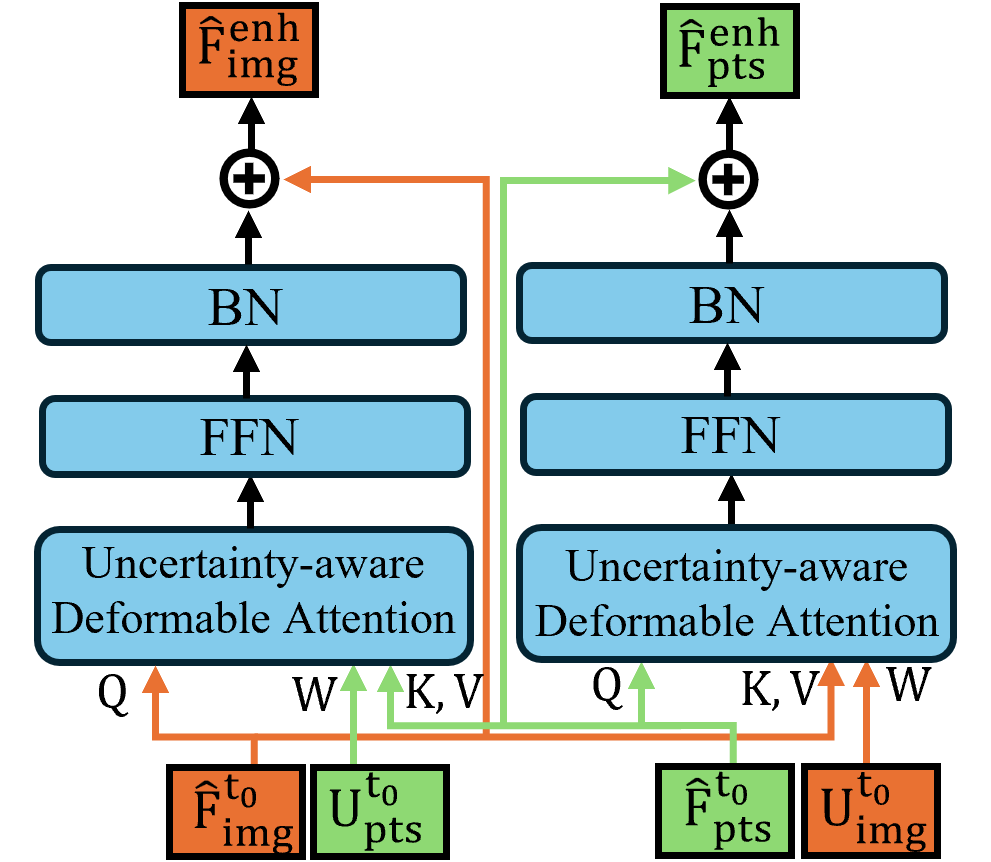}
      \caption{Uncertainty-guided cross-modality transformer, taking compensated features and uncertainty maps of two modalities as inputs to cross-modally enhance features.}
      \label{fig:CrMdFs}
   \end{figure}
   
\noindent \textbf{Cross-modality Fusion.}
Cross-modality fusion is designed to address the potential unreliability of compensated features, which may contain biases or noise from temporal prediction as well as inherent limitations of individual modalities.
By leveraging complementary information across modalities, the fusion process can suppress unreliable signals, recover missing details, and thereby enhance the robustness of the compensated features. Here, we innovate an uncertainty-aware deformable transformer block to transfer reliable complementary information from one modality to the other. Notably, on the basis of deformable attention~\cite{xia2022vision}, we incorporate the uncertainty measurement of the compensated features into attention calculation. 

Let's take the enhancement of the image feature by the LiDAR feature as an example. As shown in Fig.~\ref{fig:CrMdFs}, to ensure sensitivity to the spatial distribution of images, we use the compensated image feature $\hat{F}_{\text{img}}^{t_0}$ as the query, while the compensated LiDAR feature $\hat{F}_{\text{pts}}^{t_0}$ serves as the key and value. 
Then we can denote our LiDAR-enhanced image feature generated by our uncertainty-aware Deformable transformer block as  
\begin{equation}
    \tilde{F}_\text{img}^\text{enh} = \hat{F}_{\text{img}}^{t_0} +\mathrm{DeformAttn}_{\{U_{pts}^{t_0}\}}\big(\hat{F}_{\text{img}}^{t_0}, \hat{F}_{\text{pts}}^{t_0}\big),
\end{equation}
where the subscript $\{U_{pts}^{t_0}\}$ highlights the involvement of uncertainty values of the LiDAR feature in the deformable attention calculation. Specifically, after obtaining the base attention weights $W$ (as in standard deformable attention, predicted from the query), the contribution of the LiDAR feature is further augmented by its uncertainty map $U_{pts}^{t_0}$: 
\begin{equation}
    \tilde{W}=W \cdot [1-softmax(U_{pts}^{t_0})].
\end{equation}
This $\tilde{W}$ then replaces the original attention weights $W$ for attention calculation. Here, the softmax normalization is adopted to highlight the most unreliable components in the predicted feature.
Through a similar process, we also obtain the image-enhanced feature $\hat{F}_{\text{pts}}^\text{enh}$. 

During training, the extracted features $F_{\text{img}}^{t_0}$ and $F_{\text{pts}}^{t_0}$ from the backbone are used as supervision for the uncertainty-aware fusion, and we adopt an $\ell_2$ loss to measure the discrepancy between the fusion features and the target: 
\begin{equation}
    \mathcal{L}_{\text{Fuse}} = \lVert \hat{F}_{\text{img}}^\text{enh} - F_{\text{img}}^{t_0} \rVert_2^2 + \lVert \hat{F}_{\text{pts}}^\text{enh} - F_{\text{pts}}^{t_0} \rVert_2^2.
\end{equation}

\subsection{Training and Inference Strategy.}
\label{sec:train_test_strategy}
During training, we optimize the two modules in separate stages. The motivation is to first establish reliable temporal dynamics before introducing cross-modality interaction, thereby preventing unstable temporal predictions from propagating severe noise into the fusion stage. We denote the detector’s standard detection loss (comprising classification and regression terms) as $\mathcal{L}_\text{det}$. In the first stage, the \textit{History-based Temporal Transformer} is trained jointly with $\mathcal{L}_{\text{TemPred}}$ and $\mathcal{L}_\text{det}$ to capture temporal feature prediction: 
\begin{equation}
    \mathcal{L}_{\text{HFP}} = \mathcal{L}_{\text{TemPred}} + \mathcal{L}_\text{det}.
\end{equation}
In the second stage, we freeze the temporal transformer to stabilize the learned temporal modeling, and then optimize the \textit{Uncertainty-based Cross-modality Transformer} using $\mathcal{L}_{\text{Uncert}}$, $\mathcal{L}_{\text{Fuse}}$, and $\mathcal{L}_\text{det}$: 
\begin{equation}
    \mathcal{L}_{\text{UncertFuse}} = \mathcal{L}_{\text{Uncert}} + \mathcal{L}_{\text{Fuse}} + \mathcal{L}_\text{det}.
\end{equation}

During inference, the history memory bank is continuously updated to maintain temporal continuity under arbitrary modality drop. If a modality is available, the extracted feature $F_{M}^{t_0-1}$ is stored in the bank; otherwise, the compensated feature $\hat{F}_{M}^{t_0-1}$ is used. This mechanism prevents disruption of the temporal stream when inputs are missing. For detection, the compensated feature $\hat{F}{M}^{t_0}$ is used when a modality is absent, while in the normal case, the extracted feature $F_{M}^{t_0}$ is directly adopted.
This strategy ensures robustness by preserving temporal consistency and fully exploiting reliable features under both complete and degraded sensor conditions.

%% file: sec/4_experim.tex
\begin{table*}[t]
\centering
\caption{Performance comparison in \textit{mAP} (\%) and \textit{NDS} (\%) of detectors and their ModalPatch-enhanced counterparts on nuScenes validation set under drop\_rates = \{10\%, 30\%, 50\%\}, where ``No drop'' denotes the setting where all sensor modalities are fully available and values in parentheses denote the performance boosts over the detectors.}
\label{tab: main_detect_with_missing}
\resizebox{0.96\textwidth}{!}{%
\begin{tabular}{@{}c|cc|cc|cc|cc@{}}
\toprule
\multirow{2}{*}{\textbf{Methods}} & \multicolumn{2}{c|}{\textbf{No drop}}                  & \multicolumn{2}{c|}{\textbf{Drop\_rate = 10\%}} & \multicolumn{2}{c|}{\textbf{Drop\_rate = 30\%}} & \multicolumn{2}{c}{\textbf{Drop\_rate = 50\%}} \\ \cmidrule(l){2-9} 
                                  & \textbf{mAP}                    & \textbf{NDS}                    & \textbf{mAP}          & \textbf{NDS}          & \textbf{mAP}          & \textbf{NDS}          & \textbf{mAP}          & \textbf{NDS}          \\ \midrule
UniBEV                & \multirow{2}{*}{64.24} & \multirow{2}{*}{68.53} & 59.66        & 65.47        & 48.86        & 58.53        & 35.49        & 50.65        \\
\cellcolor{gray!20} +\textbf{ModalPatch}                     &                        &                        & \cellcolor{gray!20} 62.65 (+2.99) & \cellcolor{gray!20} 67.35 (+1.88) & \cellcolor{gray!20} 56.56 (+7.70)   & \cellcolor{gray!20} 63.40 (+4.87)  & \cellcolor{gray!20} 46.32 (+10.83) & \cellcolor{gray!20} 56.91 (+6.26)        \\ \midrule
BevFusion                & \multirow{2}{*}{68.73} & \multirow{2}{*}{71.47} & 60.74        & 67.46        & 43.26        & 58.45        & 27.74        & 50.37        \\
\cellcolor{gray!20} +\textbf{ModalPatch }                    &                        &                        & \cellcolor{gray!20} 63.37 (+2.63) & \cellcolor{gray!20} 68.43 (+0.97) & \cellcolor{gray!20} 50.59 (+7.33)  & \cellcolor{gray!20} 61.30 (+2.85)  & \cellcolor{gray!20} 31.43 (+3.69)  & \cellcolor{gray!20} 50.60 (+0.23)         \\ \midrule
CMT                      & \multirow{2}{*}{70.07} & \multirow{2}{*}{72.68} & 60.51        & 67.81        & 42.90         & 58.47        & 27.21        & 49.92        \\
\cellcolor{gray!20} +\textbf{ModalPatch}                     &                        &                        & \cellcolor{gray!20} 67.96 (+7.45) & \cellcolor{gray!20} 71.42 (+3.61) & \cellcolor{gray!20} 58.77 (+15.87) & \cellcolor{gray!20} 65.81 (+7.34) & \cellcolor{gray!20} 44.21 (+17.00)    & \cellcolor{gray!20} 56.89 (+6.97)        \\ \midrule
MEFormer                 & \multirow{2}{*}{71.49} & \multirow{2}{*}{73.86} & 62.93        & 69.40         & 45.37        & 60.29        & 27.88        & 50.64        \\
\cellcolor{gray!20} +\textbf{ModalPatch}                     &                        &                        & \cellcolor{gray!20} 68.65 (+5.72) & \cellcolor{gray!20} 72.11 (+2.71) & \cellcolor{gray!20} 59.02 (+13.65) & \cellcolor{gray!20} 66.30 (+6.01)  & \cellcolor{gray!20} 44.11 (+16.23) & \cellcolor{gray!20} 57.36 (+6.72)        \\ \bottomrule
\end{tabular}%
}
\end{table*}

\section{EXPERIMENT}
\subsection{Experiment Settings}

\noindent \textbf{Dataset and Metrics.}
We conduct experiments on the large-scale \textit{nuScenes} dataset~\cite{caesar2020nuscenes}, which contains 1000 driving scenes (850 for training and 150 for validation) collected in Boston and Singapore. The dataset is captured using a 32-beam LiDAR and six surrounding cameras covering a full $360^\circ$ field of view. We use the official training and validation splits for model training and evaluation, respectively.  
To simulate random modality-drop scenarios during testing, for each input modality $M \in \{\text{Camera}, \text{LiDAR}\}$ we apply a drop probability $\text{Drop\_rate} \in \{0.1, 0.3, 0.5\}$ to independently discard the corresponding inputs. Since modality dropping is modeled as independent events, it is possible that both modalities are absent at the same time step.  
For model evaluation, we adopt the official metrics of nuScenes~\cite{caesar2020nuscenes}, the \textit{mean Average Precision} (mAP) and \textit{nuScenes Detection Score} (NDS), as primary evaluation metrics to assess detection performance.

\noindent \textbf{Models and Implementation Details.}
Our proposed \textit{ModalPatch} is, to the best of our knowledge, the first plug-and-play module for multi-modal 3D object detection under modality-drop scenarios, which can be seamlessly integrated into existing detectors. To demonstrate its generality, we evaluate ModalPatch on two representatives of detectors: the BEV-based BEVFusion~\cite{liu2022bevfusion}, and the transformer-based CMT~\cite{yan2023cross}, together with their recent modality-missing extensions, UniBEV~\cite{wang2024unibev} and MEFormer~\cite{cha2024robust}. In all experiments, we initialize the detectors with the officially released, pre-trained weights and keep these parameters frozen during the subsequent training. 
When applying ModalPatch in training, we remove the randomness in data augmentation from each detector to ensure feature consistency across consecutive frames. For training ModalPatch, we use AdamW as the optimizer with a learning rate of 0.0002.  The training of both stages is conducted for 12 epochs. The temporal length of the memory bank is set to $\tau=6$. 
During testing, for baseline detectors without \textit{ModalPatch}, UniBEV includes a built-in mechanism to handle ``None'' inputs, and other detectors replace missing modalities with zero-filled tensors to ensure inference execution. For detectors + \textit{ModalPatch}, we follow the inference procedure described in Section~\ref{sec:train_test_strategy} to handle modality drops.
Both training and testing are conducted with a batch size of 1 on the NVIDIA RTX 3090 GPU.

\begin{figure*}[t]
      \centering
      \includegraphics[width=0.93\textwidth]{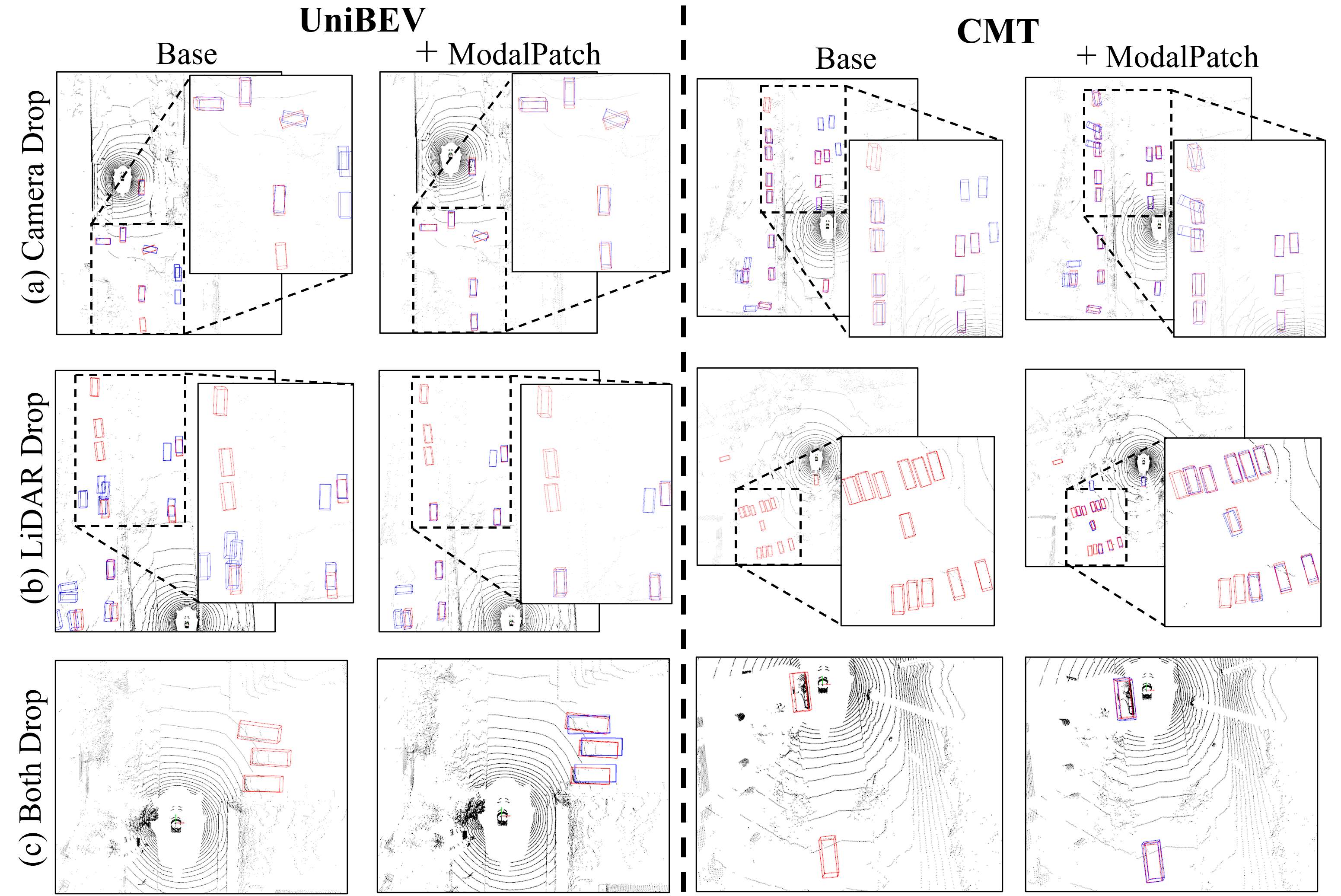}
      \caption{Qualitative visualizations of UniBEV and CMT detectors w/ or w/o ModalPatch under a 50\% drop rate. Each pair compares the baseline detector (Base) with the detector enhanced by ModalPatch (+ModalPatch), where red boxes denote ground-truth objects and blue ones denote detected objects.}
      \label{fig:detection_vis}
   \end{figure*}

\subsection{Detection under Modality Drop}

Table~\ref{tab: main_detect_with_missing} demonstrates the superiority of ModalPatch in enhancing robustness under modality-drop scenarios. While all baseline detectors suffer severe degradation, ModalPatch consistently alleviates the performance collapse across diverse architectures. 
Under a moderate drop rate of 10\%, ModalPatch provides consistent improvements across all detectors, with an average gain of +4.70\% mAP and +2.29\% NDS, effectively mitigating the performance degradation caused by missing modalities and keeping the performance closer to the no-missing upper bound.
Under the 30\% drop rate, our method yields substantial improvements (average +11.14\% in mAP and +5.28\% in NDS), showing the effectiveness of temporal feature prediction and uncertainty-guided fusion in recovering missing signals. Under the challenging 50\% drop condition, ModalPatch still achieves significant gains, leading to an average improvement of +11.93\% mAP and +5.05\% NDS. Notably, UniBEV incorporates a feature-adaptive fusion strategy that provides relatively strong baseline performance under severe modality loss; however, ModalPatch still delivers a significant margin of improvement (e.g., +10.83\% mAP / +6.26\% NDS at 50\% drop), demonstrating its complementary value. These results highlight that ModalPatch not only provides consistent benefits to both BEV-based and transformer-based detectors but also delivers the most significant advantages in extreme failure cases, confirming its practicality as a plug-and-play solution for real-world deployment. In cases of ``no drop'', ModalPatch preserves the baseline performance by leaving original features unaltered.

\noindent\textbf{Detection Visualization.} Taking UniBEV and CMT for examples, Fig.~\ref{fig:detection_vis} presents qualitative visualizations under the challenging 50\% drop rate and details three cases: camera drop, LiDAR drop, and both-drop. In the camera drop case, the baselines fail to detect many distant objects because sparse LiDAR scans alone cannot provide sufficient appearance cues. With ModalPatch, the missing image information (e.g., color and texture) is effectively compensated, leading to improved recall of far-range objects and more precise bounding boxes. In the LiDAR drop case, UniBEV exhibits box localization biases due to the lack of depth information, while CMT—being heavily dependent on LiDAR—completely misses several targets. By contrast, with ModalPatch, the compensated LiDAR features restore spatial cues, resulting in more accurate localization and higher recall. Finally, in the extreme both-drop case, neither UniBEV nor CMT can produce meaningful detections because they lack mechanisms to handle simultaneous modality loss. With ModalPatch, however, the integration of temporal prediction and uncertainty-guided cross-modal fusion still enables the detectors to capture object instances, demonstrating strong robustness and reliability under extreme sensor failures.

\begin{table}[]
\centering
\caption{Component ablation in \textit{mAP/NDS} (\%) of proposed modules at drop\_rates=50\%, where Base denotes the original detector.}
\label{tab:albation_component}
\resizebox{0.9\columnwidth}{!}{%
\begin{tabular}{@{}c|ccc|c|c@{}}
\toprule
    & \textbf{Base} & \textbf{HFP} & \textbf{UCF} & \textbf{UniBEV} & \textbf{CMT} \\ \midrule
(a) & \checkmark                 &              &                    & 35.49/50.65     & 27.21/49.92  \\
(b) & \checkmark                 & \checkmark            &                    & 42.04/54.37     & 40.45/54.85  \\ \midrule
(c) & \checkmark                & \checkmark            & \checkmark                 & 46.32/56.91     & 44.21/56.89  \\ \bottomrule
\end{tabular}%
}
\end{table}

\subsection{Ablation Studies}
We conduct ablation experiments primarily using UniBEV and CMT as the baseline detectors, under a 50\% drop rate, to further analyze the proposed modules.

\noindent \textbf{Component ablation.} Table~\ref{tab:albation_component} analyses the individual contributions of our proposed modules. Starting from the baseline detector (a), adding the history-based feature prediction module (b) significantly boosts performance (e.g., +6.55\% mAP / +3.72\% NDS on UniBEV and +13.24\% mAP / +4.93\% NDS on CMT), demonstrating that temporal feature modeling effectively compensates for missing modalities. Building on this, integrating the uncertainty-guided cross-modality fusion module (c) provides additional noticeable gains (e.g., +4.28\% mAP / +2.54\% NDS on UniBEV and +3.76\% mAP / +2.04 NDS\% on CMT), as cross-modal interaction refines the compensated features and suppresses prediction bias. Together, the two modules work complementarily, achieving robust and reliable detection even under severe modality-missing conditions.

\begin{figure}[t]
      \centering
      \includegraphics[width=0.49\textwidth]{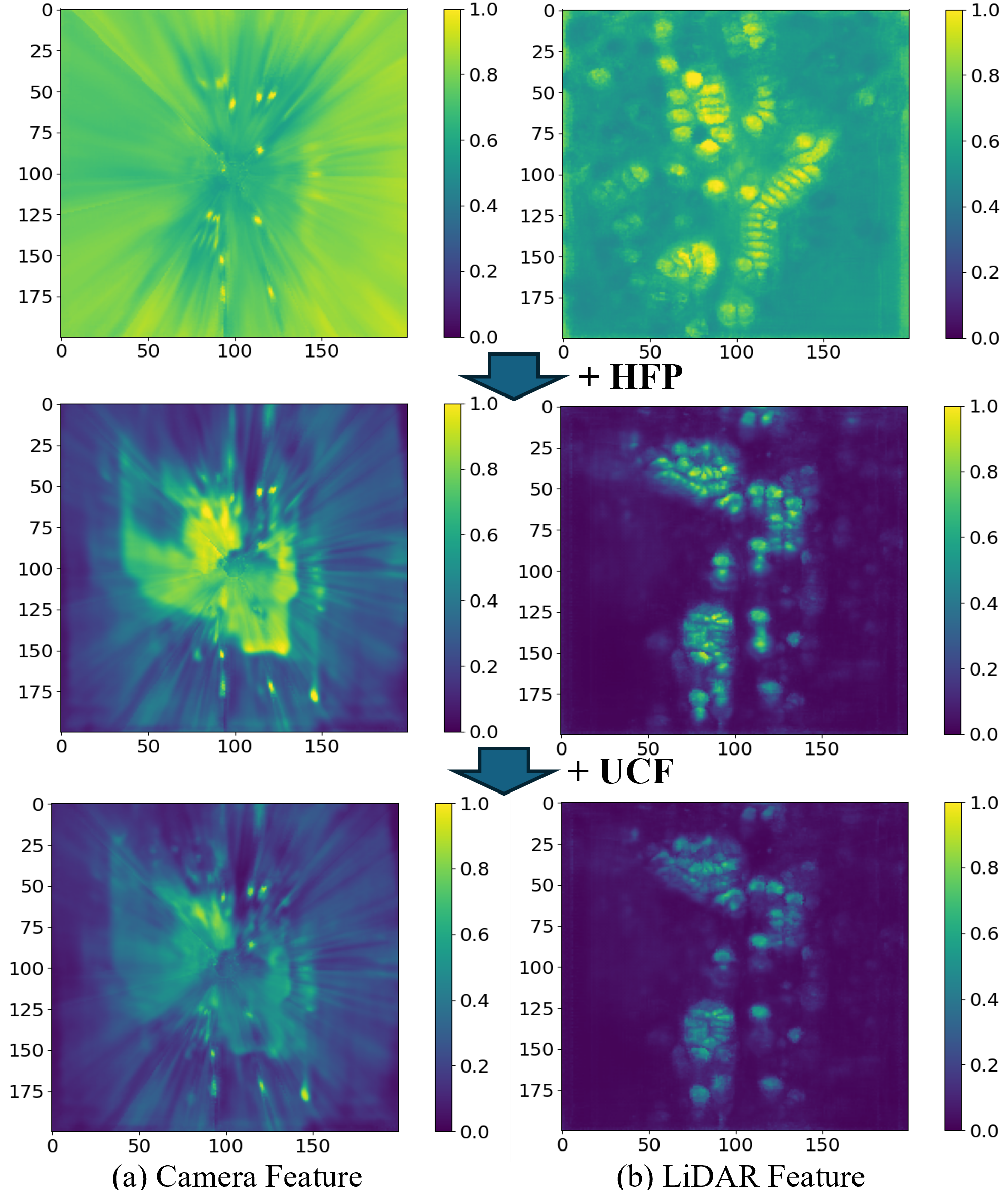}
      \caption{Feature mean squared error between the predicted features and the ground-truth features for camera (a) and LiDAR (b) modalities using UniBEV with ModalPatch, with the top row for the baseline mechanism (i.e., zero-filled feature due to modality drop), the middle row for adopting the HFP, and the bottom row for continuously adopting the UCF.}
      \label{fig:hfp_vs_crmdfs}
   \end{figure}

We also analyze the feature's \textit{mean squared error} (MSE) with respect to the ground-truth features (i.e., extracted features with no drop) for both camera and LiDAR modalities, as shown in Fig.~\ref{fig:hfp_vs_crmdfs}. The top row presents the baseline results without ModalPatch, which exhibit large reconstruction errors due to the absence of features. With the history-based feature prediction module (middle row), temporal modeling effectively reduces these errors by compensating for missing features. After further incorporating the uncertainty-guided cross-modality transformer (bottom row), the error maps are further suppressed and become more uniformly dim, indicating the effective suppression on the prediction bias. By complementarily adopting two proposed modules, the reliability of compensated features is enhanced, leading to more accurate feature recovery under missing modalities.


\noindent \textbf{Albation on uncertainty-guided cross-modality fusion.}
Table~\ref{tab:albation_uncert} compares the performance of the cross-modality transformer with and without the proposed uncertainty modeling. Without uncertainty guidance, the fusion relies purely on spatial attention, leading to moderate improvements but still propagating biased or unreliable features. By contrast, equipping the transformer with uncertainty estimation yields consistent gains (e.g., +1.39\% mAP / +1.49\% NDS on UniBEV and +1.02\% mAP / +0.92\% NDS on CMT), confirming that explicitly quantifying the reliability of compensated features allows the model to suppress biased signals and emphasize trustworthy information. 

\begin{table}[]
\centering
\caption{Ablation of the proposed UCF with and without uncertainty modeling in \textit{mAP} (\%) and \textit{NDS} (\%) under 50\% modality drop.}
\label{tab:albation_uncert}
\resizebox{0.95\columnwidth}{!}{%
\begin{tabular}{@{}c|cc|cc@{}}
\toprule
\multirow{2}{*}{\textbf{Methods}} & \multicolumn{2}{c|}{\textbf{UniBEV}} & \multicolumn{2}{c}{\textbf{CMT}} \\ \cmidrule(l){2-5} 
                  & \textbf{mAP}     & \textbf{NDS}     & \textbf{mAP}    & \textbf{NDS}   \\ \midrule
w/o uncertainty modeling        & 44.93            & 55.42            & 43.19           & 55.97          \\
w/ uncertainty modeling        & 46.32            & 56.91            & 44.21           & 56.89  \\ \bottomrule       
\end{tabular}%
}
\end{table}

\begin{table}[]
\centering
\caption{Results in \textit{mAP} (\%) and \textit{NDS} (\%) under the 50\% drop rate scenario when only a single modality (LiDAR-only or Camera-only) is available. Since no complementary modalities can be utilized, only the HFP module is adopted for temporal compensation.}
\label{tab:ablation_single_modal}
\resizebox{0.87\columnwidth}{!}{%
\begin{tabular}{@{}c|cc|cc@{}}
\toprule
\multirow{2}{*}{\textbf{Methods}} & \multicolumn{2}{c}{\textbf{LiDAR-only}} & \multicolumn{2}{c}{\textbf{Camera-only}} \\ \cmidrule(l){2-5} 
                         & \textbf{mAP}   & \textbf{NDS}  & \textbf{mAP}   & \textbf{NDS}  \\ \midrule
UniBEV                   & 23.85          & 47.58         & 13.45          & 31.07         \\
+HFP                     & 36.62          & 52.50         & 24.77          & 36.48         \\ \midrule
No modality drop               & 58.16          & 65.26         & 35.00          & 42.40         \\ \midrule \midrule
CMT                      & 24.13          & 48.37         & 0.02           & 1.69          \\
+HFP                     & 27.82          & 49.78         & 0.17           & 5.03          \\ \midrule
No modality drop               & 60.01          & 67.23         & 0.26           & 8.84          \\ \bottomrule
\end{tabular}%
}
\end{table}

\noindent \textbf{Ablation on single-modality conditions.}
Table~\ref{tab:ablation_single_modal} evaluates the performance under the 50\% drop rate when only a single modality (camera or LiDAR) is available. Since no complementary modality can be exploited in this setting, we only employ the HFP module to provide temporal compensation. The results show that HFP brings substantial improvements for UniBEV, with gains of +12.77\% mAP / +4.92\% NDS in the LiDAR-only case and +11.32\% mAP / +5.41\% NDS in the camera-only case, demonstrating its strong effectiveness in alleviating single-modality collapse. For CMT, HFP also improves the LiDAR-only performance by +3.69\% mAP / +1.41\% NDS. In the camera-only case, the gain is more modest, as the detector shows a strong reliance on LiDAR and its performance ceiling with no camera input missing is inherently low (0.26\% mAP / 8.84\% NDS). These results demonstrate that our method continues to yield measurable improvements, suggesting that temporal modeling can enhance robustness and preserve detection capability even with only a single modality existing.



\begin{table}[t]
\centering
\caption{Runtime speed measured with frames per second (FPS) under Drop\_rate=50\% on NVIDIA RTX 3090 GPU.}
\label{tab:ablation_speed}
\resizebox{0.98\columnwidth}{!}{%
\begin{tabular}{@{}c|cccc|c@{}}
\toprule
\textbf{Methods} & \textbf{UniBEV} & \textbf{BEVFusion} & \textbf{CMT} & \textbf{MEFormer} & \textbf{Ave} \\ \midrule
Base             & 3.21            & 8.03               & 4.53         & 5.56 & 5.33              \\
+ModalPatch      & 3.06            & 7.54               & 4.37         & 4.62 & 4.90             \\ \bottomrule
\end{tabular}%
}
\end{table}

\noindent\textbf{Runtime Speed.}
Table~\ref{tab:ablation_speed} shows the runtime speed measured in frames per second (FPS) with a 50\% drop rate. On average, integrating ModalPatch results in an average runtime reduction from 5.33 FPS to 4.90 FPS, while providing substantial robustness and accuracy improvements under modality-missing conditions as in Table~\ref{tab: main_detect_with_missing}. This demonstrates that ModalPatch achieves a favorable trade-off between detection performance and computational efficiency, making it practical for real-world deployments with modality drop.

\subsection{Limitation}
Despite the effectiveness of ModalPatch in improving robustness under modality-drop scenarios, its benefits are constrained when the detector performs poorly on a single modality with inherently low detection capability, as our method primarily compensates for the unfavorable modality's features through history-based feature prediction. In the future, we plan to explore stronger single-modality feature enhancement to further overcome this limitation.

%% file: sec/5_conclude.tex
\section{CONCLUSION}

In this paper, we present ModalPatch, a plug-and-play robustness module for multi-modal 3D object detection, which aims to enable existing detectors to cope with independent sensor drops without requiring significant retraining or architectural modifications. ModalPatch consists of two complementary components: a history-based feature prediction module that leverages temporal continuity to model the evolution of features across consecutive frames and predict current representations for missing inputs; and an uncertainty-guided cross-modality fusion module that estimates the reliability of the compensated features and performs spatially-aware cross-modal fusion, suppressing biased signals while reinforcing trustworthy information. Extensive experiments demonstrate that ModalPatch consistently improves the robustness of diverse detectors under various modality-drop conditions.